\documentclass[10pt,twocolumn,letterpaper]{article}

\usepackage{wacv}              

\usepackage{graphicx}
\usepackage{amsmath}
\usepackage{amssymb}
\usepackage{booktabs}
\usepackage{times}
\usepackage{latexsym}
\usepackage[T1]{fontenc}
\usepackage{graphicx}
\usepackage[utf8]{inputenc}
\usepackage{microtype}
\usepackage{inconsolata}
\usepackage[accsupp]{axessibility}

\usepackage[pagebackref,breaklinks,colorlinks]{hyperref}

\usepackage[capitalize]{cleveref}
\crefname{section}{Sec.}{Secs.}
\Crefname{section}{Section}{Sections}
\Crefname{table}{Table}{Tables}
\crefname{table}{Tab.}{Tabs.}

\def\wacvPaperID{3} 
\def\confName{WVLL}
\def\confYear{2024}

\usepackage{latexsym}
\usepackage{scalerel}
\usepackage{tikz}
\usetikzlibrary{svg.path}

\definecolor{orcidlogocol}{HTML}{A6CE39}
\tikzset{
  orcidlogo/.pic={
    \fill[orcidlogocol] svg{M256,128c0,70.7-57.3,128-128,128C57.3,256,0,198.7,0,128C0,57.3,57.3,0,128,0C198.7,0,256,57.3,256,128z};
    \fill[white] svg{M86.3,186.2H70.9V79.1h15.4v48.4V186.2z}
                 svg{M108.9,79.1h41.6c39.6,0,57,28.3,57,53.6c0,27.5-21.5,53.6-56.8,53.6h-41.8V79.1z M124.3,172.4h24.5c34.9,0,42.9-26.5,42.9-39.7c0-21.5-13.7-39.7-43.7-39.7h-23.7V172.4z}
                 svg{M88.7,56.8c0,5.5-4.5,10.1-10.1,10.1c-5.6,0-10.1-4.6-10.1-10.1c0-5.6,4.5-10.1,10.1-10.1C84.2,46.7,88.7,51.3,88.7,56.8z};
  }
}

\newcommand\orcidicon[1]{\href{https://orcid.org/#1}{\mbox{\scalerel*{
\begin{tikzpicture}[yscale=-1,transform shape]
\pic{orcidlogo};
\end{tikzpicture}
}{|}}}}

\begin{document}

\title{Enhancement of Bengali OCR by Specialized Models and Advanced Techniques for Diverse Document Types}

\author{
AKM Shahariar Azad Rabby\textsuperscript{2,3 \orcidicon{0000-0003-3994-3105}}, 
Hasmot Ali\textsuperscript{2 \orcidicon{0000-0002-8885-2501}}, 
Md. Majedul Islam\textsuperscript{2 \orcidicon{0000-0002-9318-056X}}, 
Sheikh Abujar\textsuperscript{3 \orcidicon{}}, 
Fuad Rahman\textsuperscript{1 \orcidicon{0000-0002-8670-7124}}\\
\textsuperscript{1}Apurba Technologies, CA, USA\\
\textsuperscript{2}Apurba Technologies, Dhaka, Bangladesh\\  
\textsuperscript{3}The University of Alabama at Birmingham, AL, USA\\
\small \{rabby, majed, fuad\}@apurbatech.com, hasmot\_ali@apurba.com.bd, \{arabby, sabujar\}@uab.edu
}

\maketitle

\begin{abstract}
This research paper presents a unique Bengali OCR system with some capabilities. The system excels in reconstructing document layouts while preserving structure, alignment, and images. It incorporates advanced image and signature detection for accurate extraction. Specialized models for word segmentation cater to diverse document types, including computer-composed, letterpress, typewriter, and handwritten documents. The system handles static and dynamic handwritten inputs, recognizing various writing styles. Furthermore, it has the ability to recognize compound characters in Bengali. Extensive data collection efforts provide a diverse corpus, while advanced technical components optimize character and word recognition. Additional contributions include image, logo, signature and table recognition, perspective correction, layout reconstruction, and a queuing module for efficient and scalable processing. The system demonstrates outstanding performance in efficient and accurate text extraction and analysis.
\end{abstract}


\section{Introduction}
The advancement of optical character recognition (OCR) systems has revolutionized the digitization and analysis of textual content. The presented approach is designed to accurately reconstruct document layouts while preserving the original structure and alignment of paragraphs, tables, and numbered lists. Additionally, it goes beyond layout reconstruction by restoring embedded images, ensuring a faithful representation of the original content. Another feature is incorporating advanced image and signature detection, enabling accurate identification and extraction of these elements from diverse documents.

Our research paper presents a comprehensive Bengali OCR system that addresses the challenges of processing diverse document types. Our system incorporates specialized models for word segmentation, catering to specific document types like computer-composed, letterpress, typewriter, and handwritten documents. It demonstrates versatility by accurately handling dynamic handwritten inputs and excels in recognizing compound characters used in the Bengali language, ensuring precise character recognition.

We collected a large and diverse image corpus for Bengali OCR development to build a robust foundation. The corpus includes computer-composed, typewriter, letterpress documents, and offline and online handwritten Bengali words and characters. The data collection process focused on achieving a balanced representation across various factors. Each data point in the corpus underwent meticulous human annotation, resulting in a gold-standard dataset. We also incorporated dynamic handwritten word data, enhancing the diversity and accuracy of our OCR system.

The technical components of our OCR system maximize accuracy and efficiency. We employ advanced techniques such as automatic perspective correction and character segmentation to improve recognition performance. Our previous work \cite{charactermodel}, self-attentional VGG-based multi-headed Neural Network architecture leverages self-attention mechanisms to capture intricate dependencies within characters. Post-processing techniques intelligently merge recognized characters into words, optimizing word formation and improving text readability.

We constructed a dataset with varied font sizes, families, and backgrounds for word recognition. Real-world fine-tuning and model quantization techniques were used to enhance performance and optimize resource usage. Our rule-based OCR layout module achieves accurate document layout reconstruction, distinguishing between paragraph and table regions. It reproduces numbered list items, restores images, and excels in reconstructing tables, preserving alignment and structure.

The Bengali OCR system presents a comprehensive solution for text extraction and analysis. Our system achieves high accuracy and efficiency with robust data collection, specialized models, advanced techniques, and efficient processing. The technical components and methodologies employed optimize resource usage, ensure multi-platform compatibility and enhance scalability. These advancements position our Bengali OCR system as a robust and innovative solution for various text extraction and analysis tasks.

The main contributions of this paper are as follows:
\begin{itemize}
    \item We present a Bengali OCR system that can accurately recognize text from a wide range of documents, including computer-composed, letterpress, typewriter, and handwritten documents.
    \item We develop specialized word and character segmentation models tailored to different document types.
    \item We propose a layout reconstruction module that can accurately reconstruct the original structure and layout of documents.
    \item We implement a queuing module that facilitates an asynchronous pipeline, improving the efficiency and scalability of the system.
\end{itemize}

\section{Literature Review}
Recent years have witnessed amazing developments in the area of optical character recognition thanks to advances in deep learning and neural network algorithms, advanced data processing methods, improved layout analysis tools, and multi-language support. Due to these advancements, OCR technology is now widely used for document analysis, such as text type extraction by Majedul et al.\cite{islam2021text}, data digitization like text generation by Sadeka et al. \cite{haque2018onkogan}, and information extraction \cite{omee2012complete, pramanik2018shape, rabby2018universal} which includes recognizing handwriting text extraction even if the handwriting is in complex shape. Notable OCR engines include Tesseract by Smith \cite{TessOverview} and OCRopus by Breuel \cite{Breuel2008TheOO}, which are widely used for OCR tasks in English. However, developing cutting-edge OCR technology for languages other than English is still a work in progress. Some notable examples include OCR for Chinese by Huo \cite{china} where the author shows a high-performance OCR for machine-printed documents and OCR for Arabic by Cheung et al. \cite{arab} where they applied a recognition-based segmentation approach. Even as technology advances, OCR is still vital for providing effective document processing and analysis.

In recent years, there have been notable efforts in the field of Bangla OCR by various researchers \cite{rabby2018ekush, alam2010complete, bag2013survey, silberpfennig2015improving, jannatul2021matrivasha} by providing Bangla characters, words and digits dataset for printed documents as well as handwritten documents, surveying the improvement of Bangla ocr, improving the performance of existing ocr, presenting different domain based ocr like ocr for printed characters, handwritten words, etc. Ahmed et al.\cite{ban1} proposed an effective Bangla OCR system by introducing a new recognition system for scripted Bangla characters. In contrast, Hasan et al.\cite{ban2} presented an intelligent OCR approach that could recognize printed characters from both Bangla and English. A significant challenge in Bangla OCR is the coverage of various document types. Islam et al.\cite{lay1} contributed to this area by addressing three types of Bangla documents, including computer compose, letterpress, and typewritten text, for character recognition. Layout understanding is a vital aspect of advanced OCR systems. Qiao et al.\cite{lay2} focused on English OCR with multi-modal document understanding, analyzing the layouts of different document types. Mindee \cite{doctr} introduced an OCR system that understands document layouts and can reconstruct them, providing advanced layout analysis capabilities that enhance OCR technology's applicability across multiple domains. For complex layout analysis, such as newspapers, Jhu et al.\cite{docbed} presented a multi-stage OCR approach tailored to documents with intricate layouts.

Moreover, OCR systems have been developed to recognize tables, logos, signatures, and images within documents. Rausch et al.\cite{docperse} contributed to parsing hierarchical document structures from rendering, while Chi et al.\cite{table1} proposed an approach for extracting information from documents with complex table structures. The capabilities of OCR technology have increased due to these developments in identifying and processing different document elements.

Our OCR system, which is effective at identifying images, signatures, and logos within documents, is presented in this work. It can precisely identify and reconstruct tables and has perspective correction capabilities. Our suggested OCR technology is an advanced method for extracting text in Bangla and enabling the digitalization of information.

\section{Proposed Methodology}

\subsection{Data Collection}
\begin{figure}[!ht]
\centering
\includegraphics[width=0.90\linewidth]{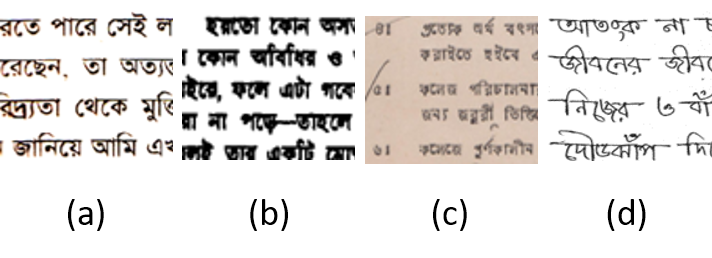}
\caption{Data Sample- a) Computer-composed b) Letterpress c) Typewritten d) Handwritten.}
\label{sampleDoc}
\end{figure}

To develop this Bengali OCR system, we undertook a comprehensive data collection process, resulting in the most extensive image corpus specifically tailored for Bengali OCR development. This corpus encompasses diverse data types, including computer-composed, typewriter, and letterpress documents, as well as offline and online handwritten Bengali words and characters. Figure \ref{sampleDoc} shows the sample data of Computer Compose, Letterpress, Typewriter, and Handwritten documents.

\begin{figure}[htbp]
\centerline{\includegraphics[height=2cm]{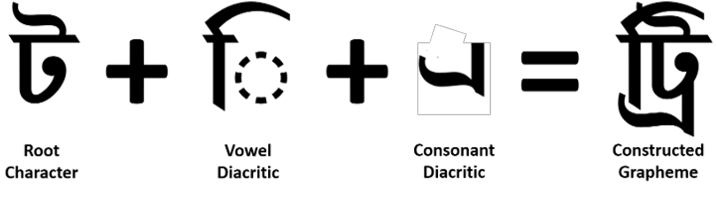}}
\caption{Bangla text structure.}
\label{figCh}
\end{figure}

The structure of the Bangla language is quite complex, which is another reason for the unsatisfactory performance of Bangla OCR. Figure \ref{figCh} visualizes the complex text structure of Bangla language.

To ensure the highest quality data, we assembled a substantial research community consisting of over 100 document processing Engineers responsible for segmentation and annotation of the data and 5 Research Assistants for supervision. The handwritten data was collected from various regions across Bangladesh, with careful attention given to a balanced representation of gender, age, occupation, and geographical factors. Each data point in our corpus underwent meticulous human annotation involving three annotators and a supervisor, creating a gold-standard dataset. Additionally, we acquired a significant amount of dynamic handwritten word data, totaling 1 million words, which was captured using a graphics tablet and included precise coordinate values.

\begin{figure}[htbp]
\centerline{\includegraphics[height=3.7cm]{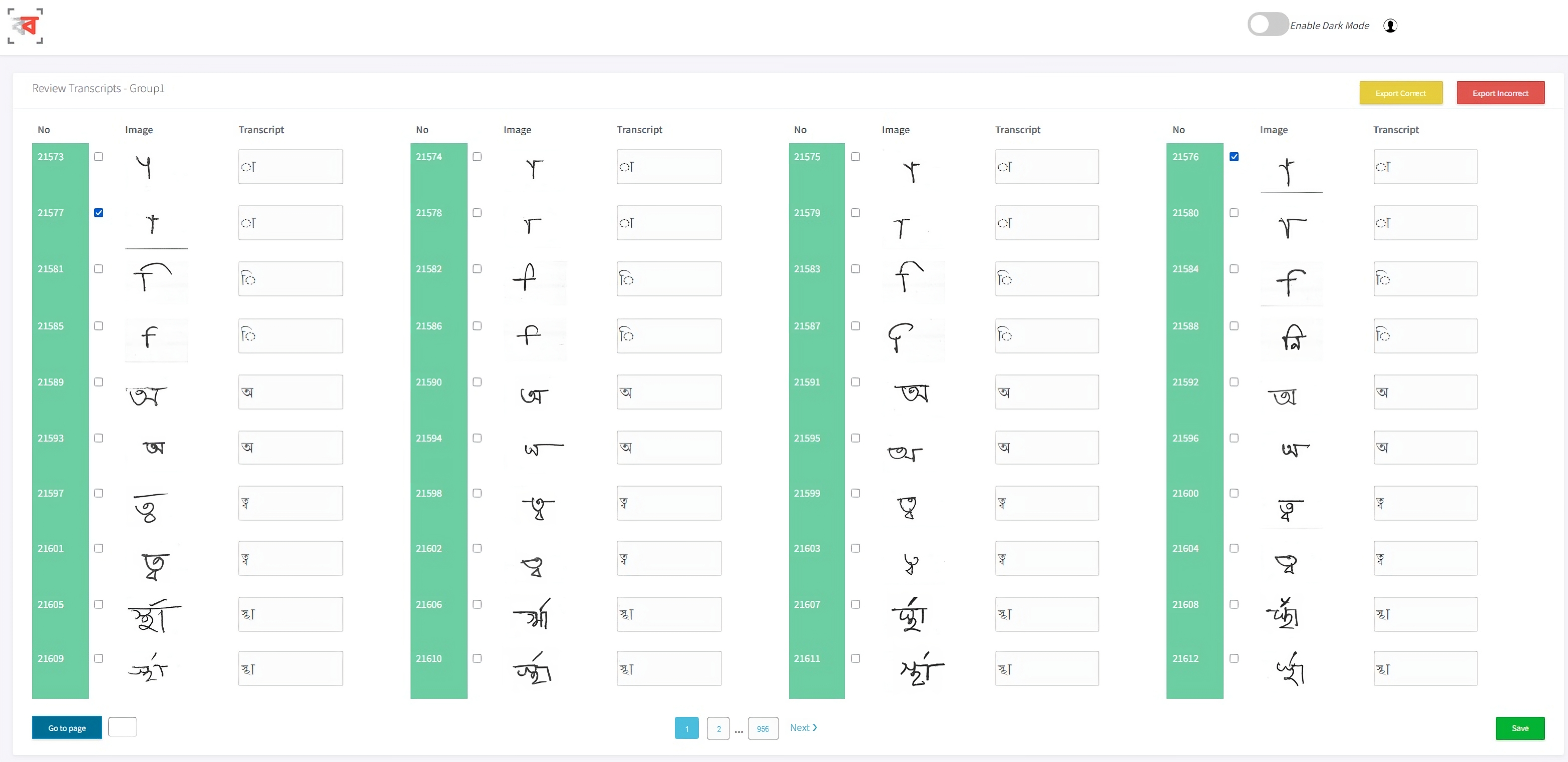}}
\caption{Annotation Tools Interface for Normal Annotator.}
\label{fig3}
\end{figure}

\begin{figure}[htbp]
\centerline{\includegraphics[height=3.7cm]{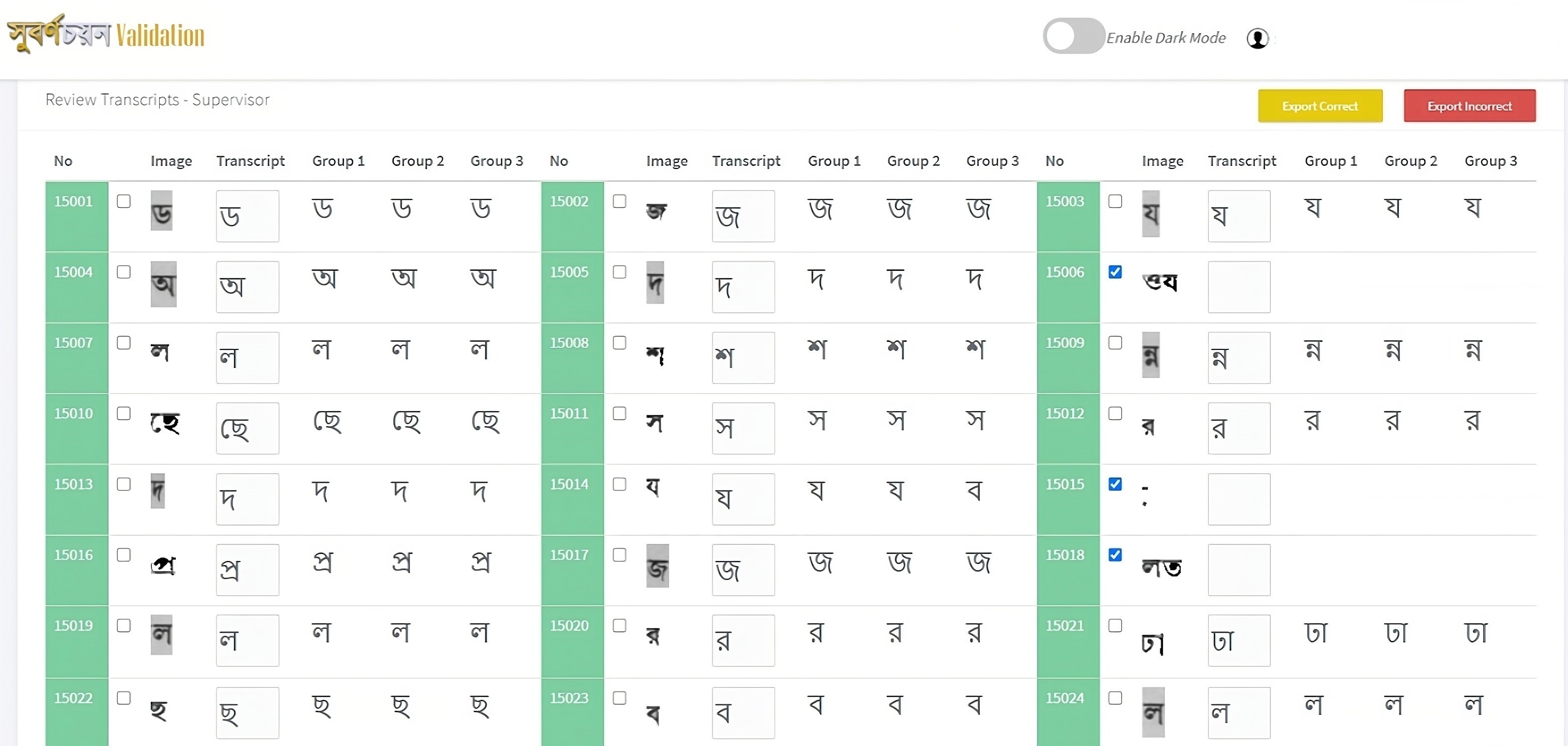}}
\caption{Annotation Tools Interface for Supervisor-level Annotator.}
\label{fig4}
\end{figure}

The collected data was annotated using our annotation application and tool. We mapped each character and diacritic to a unique label, allowing the model to use the mapping as a label for training. The annotation process involved a detailed description and analysis of the dataset, including metadata and tagging. We implemented an Android app for the annotation module, where documents were segmented into character levels, and an initial OCR model predicted character outputs. A group of annotators verified these outputs and corresponding images. Correct predictions were used for the character image corpus, while the annotators manually annotated incorrect predictions. The annotation interface displayed the character image and its corresponding text on the same screen, allowing the annotators to make accurate annotations. The annotators had options to mark the character with a textbox and select three properties (bold, italic, underline) using checkboxes. Annotators could skip character images incorrectly segmented or containing excessive noise. Disagreed images were sent to resolvers, who acted as supervisors to resolve conflicts among annotators and provide the final decision. Figure \ref{fig3},\ref{fig4}
shows the user interface of the annotation tools for normal
annotators and supervisor-level annotation accordingly.

\begin{figure}[!ht]
\centering
\includegraphics[width=\linewidth]{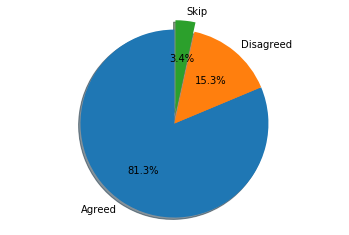}
\caption{Annotators’ Decision Ratio After the Data Annotation.}
\label{ratio}
\end{figure}

After completing the annotation process, we generated statistics and analysis for the annotated documents. The analysis revealed the agreement percentages for computer-composed, letterpress, and typewriter characters, which were 91.8\%, 80.3\%, and 64.2\%, respectively. We also analyzed the agreed data's presence and absence of various graphemes, vowel diacritics, and consonant diacritics. Furthermore, we examined the annotators' profiles and found that 81.3\% of the data had an agreement among the annotators, while 15.3\% disagreed and 3.4\% were skipped. Figure  \ref{ratio} visualize the insight. 

These analyses provide insights into the annotators' contributions and the overall quality of the annotated data. The final dataset consists of 720754 Computer Compose, 251332 Typewriter, 84295 Letterpress, and 1594924 Handwriting images. Table \ref{table3} shows the number of data before and after annotation.

\begin{table}[htbp]
\caption{Number of data before and after annotation}
\begin{center}

\begin{tabular}{|p{3cm}|p{2cm}|p{2cm}|}
\hline
\textbf{Document Type} & \textbf{Before Annotation} & \textbf{After Annotation}\\
\hline
Computer compose & 837379 & 720754\\
\hline
Letterpress& 1104304 &496543\\
\hline
Typewriter & 773043 & 251332\\
\hline
Handwritten & 1674270 & 1594924\\
\hline
\end{tabular}
\label{table3}
\end{center}
\end{table}

The data collection process and annotation were crucial for building our Bengali OCR system. We collected diverse data, balanced across different document types, domains, and demographics. The annotation application and tool ensured the accurate labeling of characters and diacritics, facilitating the training of our OCR models. The resulting dataset, detailed analysis, and statistics serve as a valuable resource for training and evaluating our Bengali OCR system.

\subsection{Model Description and Training}
Our OCR system incorporates advanced techniques and methodologies to achieve accurate and efficient character and word recognition. We begin with automatic perspective correction, eliminating distortions in document images, resulting in improved image quality for further processing. A key aspect of character recognition is the character segmentation model, which extracts individual characters from words, ensuring precise recognition and reducing the risk of misclassifications. By segmenting characters, our system can focus on recognizing each component accurately, leading to higher overall recognition accuracy.

For character recognition, our previous work \cite{charactermodel}, which is a self-attentional VGG-based multi-headed Neural Network, is used. This architecture leverages self-attention mechanisms to capture intricate dependencies and contextual information within characters, and the VGG-based backbone provides strong feature extraction capabilities, enabling the model to learn discriminative representations of characters. After character recognition, the system intelligently merges the recognized characters back into words using post-processing techniques, eliminating unnecessary spaces and ensuring the output words closely resemble the original word structures.
\begin{figure*}[htbp]
\centering
\includegraphics[width=0.6\linewidth, height=10cm]{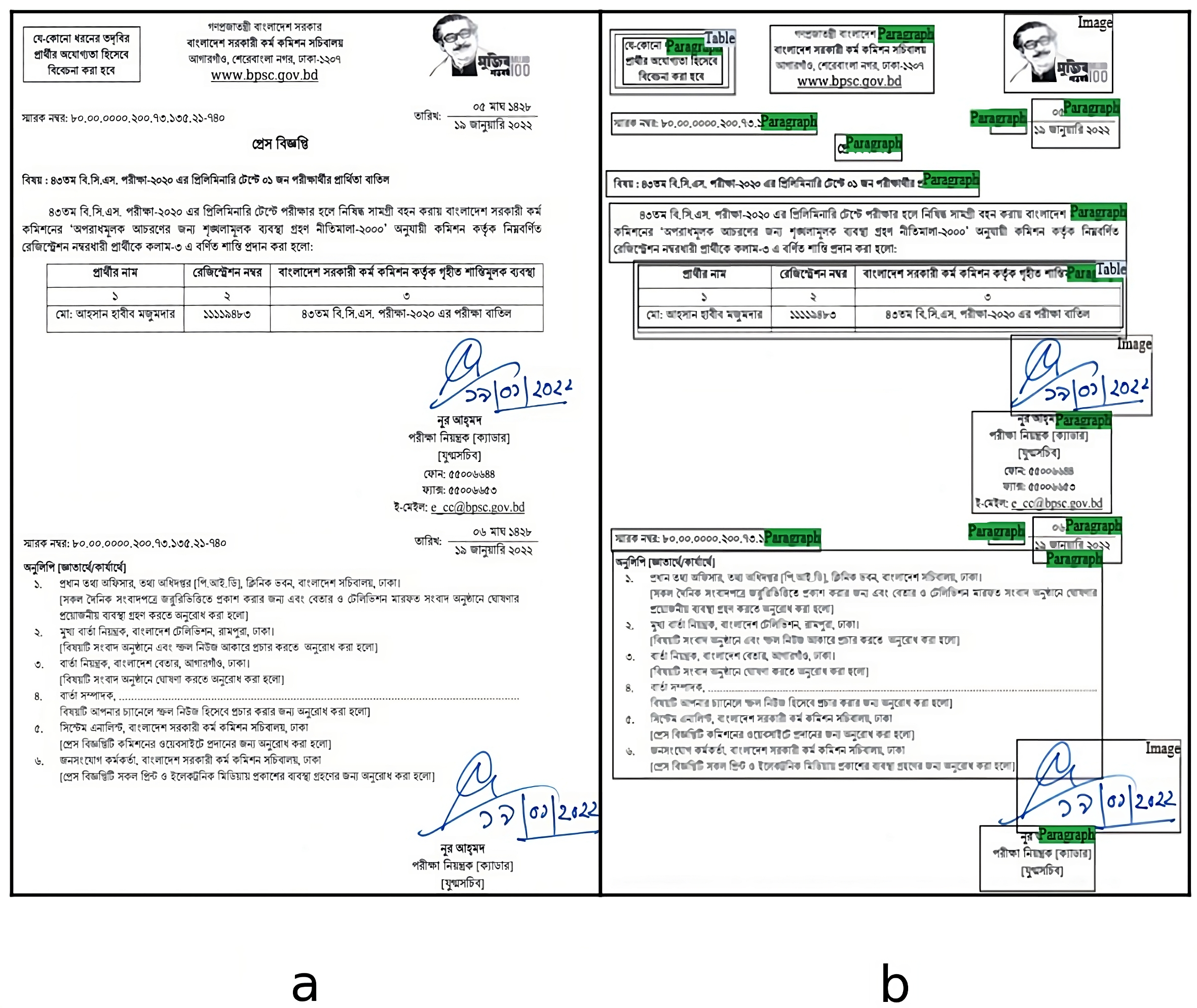}
\caption{(a) Input Image (b) Layout Analysis}
\label{fig1}
\end{figure*}
To enhance word recognition, a diverse dataset enabled us to train our models to be robust and adaptable to different document types, including computer-composed, letterpress, typewriter, and handwritten documents. To ensure the efficacy of our models in real-world scenarios, we fine-tuned them using additional real-world data. To optimize performance on CPU-based systems, we employed model quantization for converting the float weight into 8-bit integers using Quantization Aware Training, which was performed on the feature extractor backbone to learn the quantization parameters and Dynamic Quantization which was performed on the fully connected layers, resulted in the reduction of memory footprint. The pixel values of the images were normalized using the Minmax normalizer. We employed the Adam optimizer and categorical cross-entropy loss function. The model was trained on a Tesla V100-SXM2 GPU and converged in approximately 13 hours.

In some instances, rule-based approaches were incorporated alongside machine learning algorithms to improve detection performance. Rule-based techniques were particularly effective for handling commonly used punctuations, allowing for more time-efficient solutions and improved detection. Through the combination of advanced model architectures, data augmentation, fine-tuning, model quantization, and optimization for CPU deployment, our OCR system achieves high-performance character and word recognition across various document types. Using rule-based techniques complements the machine learning algorithms, ensuring reliable and efficient detection.

\subsection{Layout Detection and Reconstruction}
Layout detection and reconstruction is a critical component of our OCR system, and we have implemented a robust rule-based system to identify and reconstruct various layout elements accurately. The system effectively distinguishes between paragraph and table regions, allowing us to reconstruct lines and maintain text alignment within paragraphs, thus preserving the original layout structure. Our layout module goes beyond paragraph reconstruction and can accurately reproduce numbered list items, ensuring that the structure and formatting of numbered lists are faithfully preserved. Additionally, the module is equipped to restore images within the document, maintaining their original placement and ensuring a visually accurate representation.

Rule-based OCR layout module accurately reconstructs document layout, distinguishing between paragraph and table regions. It reproduces numbered list items, restores images, and excels in reconstructing tables, preserving alignment and structure. The rules include text region detection, possible text line height calculations, image detection, text line generation, possible text line gap calculations, possible rectangle gap calculation, paragraph region detection, page column layout generation, table detection(bordered table, semi-bordered table, and borderless table).

One of the significant challenges in layout reconstruction lies in accurately reconstructing tables, and our OCR layout module excels in this area. It achieves perfect table reconstruction by correctly aligning the text within each cell, thereby preserving the integrity of the table structure. This capability is particularly valuable when tables are crucial for data representation.

By incorporating this rule-based system into our OCR layout module, we enhance the overall accuracy and fidelity of the reconstructed document layout. The module's ability to accurately reconstruct paragraphs, lists, tables, and images provides a comprehensive solution for capturing the original structure and layout of the processed documents. This ensures that the output documents maintain the exact visual representation as the original, enabling users to work with the content familiarly and intuitively.

Figure~\ref{fig1} shows the original input image with the segmented layout structure.

\begin{figure*}[htbp]
\centering
\includegraphics[width=0.6\linewidth, height=10cm]{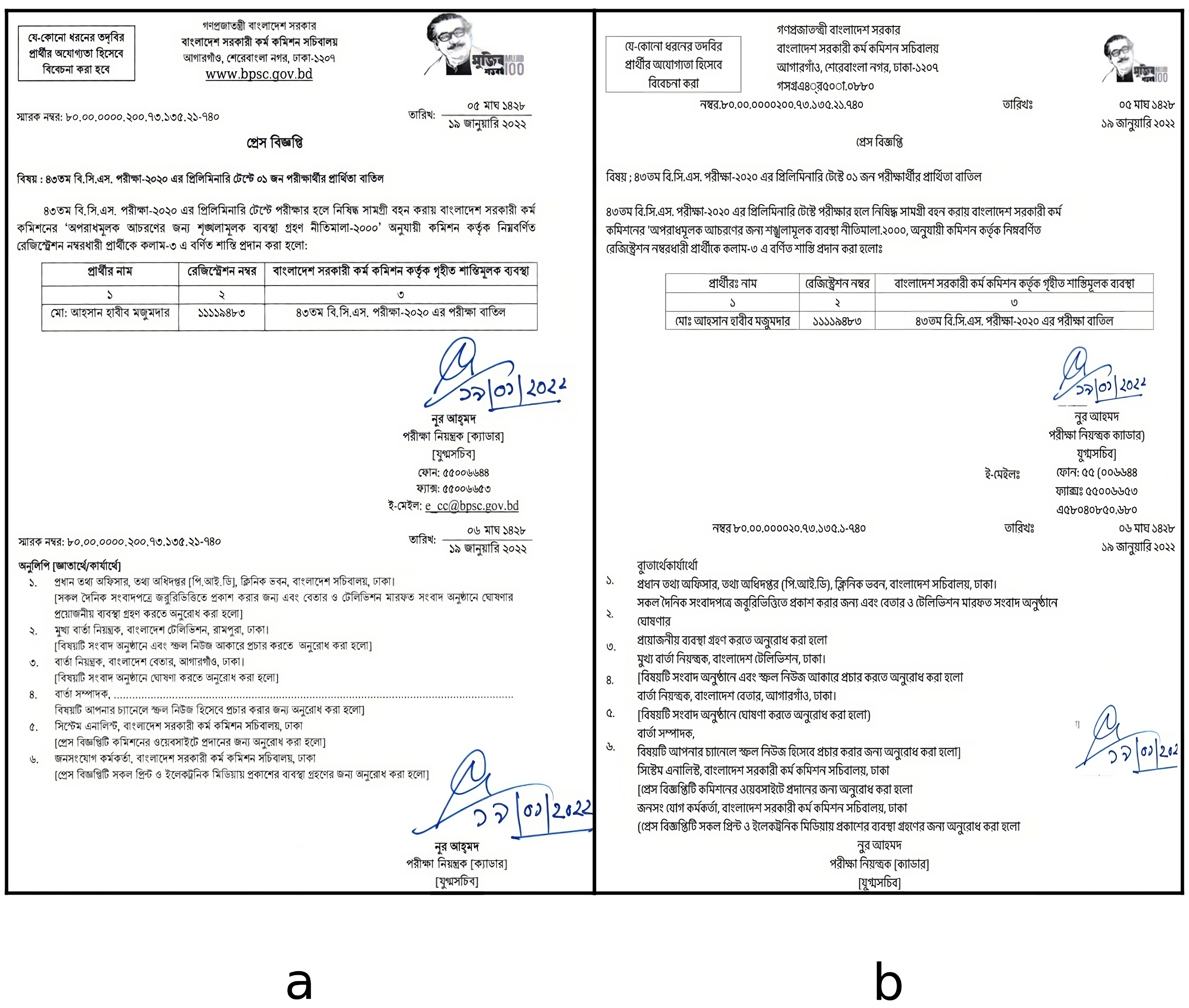}
\caption{(a) Input Image (b) Layout Analysis}
\label{fig2}
\end{figure*}

\subsection{Deployment and Scalability }
Deployment and scalability are crucial aspects of our OCR system, and we have implemented several strategies to enhance efficiency and handle large workloads. Firstly, we have incorporated a queuing module using Apache Kafka and Zookeeper, enabling an asynchronous pipeline for image processing. This queuing system ensures a smooth and uninterrupted workflow by seamlessly managing the flow of images through various processing steps. Each image is queued and processed asynchronously, eliminating the need for synchronous requests and improving overall processing speed.

Based on Apache Kafka \cite{kafka}, and Zookeeper \cite{zookeeperoverview}, the queuing module enables smooth image processing, reduces timeouts, and efficiently handles large files. The queuing system offers significant advantages, including reducing the possibility of request timeouts and facilitating the processing of new files while previous files are still being processed. This parallel processing capability improves throughput and allows for faster turnaround times. Moreover, the queuing module effectively handles large PDF files by converting every page to a single image, preventing performance bottlenecks and ensuring efficient processing.

To ensure multi-platform compatibility, we have converted our trained models to the ONNX \cite{onnx}  format, facilitating seamless integration and deployment across different deep-learning frameworks and platforms. Additionally, we have leveraged the Triton \cite{triton}  framework for efficient model deployment and scaling in production environments. Triton provides a robust infrastructure that allows us to handle large workloads and ensures reliable performance. Our OCR system takes an average time of 1.847 seconds for computer-compose, 3.319 seconds for handwritten, 1.756 seconds for typewriters, and 3.276 seconds for letterpress documents.

\section{Model Performance}
The performance of our OCR model was evaluated using randomly selected documents from each document type, which is entirely unseen when the character and word model are trained. These documents varied in font types, sizes, noise levels, and sources. Each document was processed through the inference pipeline, which involved noise removal and word boundary detection using docTR \cite{doctr} for character segmentation. The segmented characters were then passed to the model for grapheme prediction, words, and sentence generation. 
Figure ~\ref{fig2} shows the original input image with the constructed output with the proper layout.

The accuracy of the OCR output was measured by comparing it with the corresponding ground truth using the Levenshtein distance \cite{laven} and confusion matrix.

The Confusion matrix-based and Levenshtein distance-based accuracy results for each document type are presented in Table~\ref{table1}. The Confusion matrix-based accuracy reflects the overall performance considering noise, addition, deletion, and substitution errors. In contrast, the Levenshtein distance-based accuracy measures the similarity between the OCR output and the ground truth. 

\begin{table}[htbp]
\caption{Confusion matrix-based and Levenshtein distance-based accuracy}
\begin{center}

\begin{tabular}{|p{3cm}|p{1.5cm}|p{1.8cm}|}
\hline
\textbf{Document Type} & \textbf{Confusion matrix} & \textbf{Levenshtein distance}\\
\hline
Computer compose & 99.56\% & 90.06\%\\
\hline
Letterpress& 99.33\% &88.53\%\\
\hline
Typewriter & 97.98\% & 83.38\%\\
\hline
Handwritten & 95.32\% & 86.84\%\\
\hline
Average& 98.05\% & 87.20\%\\
\hline
\end{tabular}
\label{table1}
\end{center}
\end{table}
Confusion Matrix-based words lexical variation analysis presented by R. Martin \cite{matrix} for ocr performance is a traditional ocr evaluation method. The approach is also effective for evaluating the document layout analysis as presented by Alberti et al. \cite{Alberti}. On average, our system achieved a confusion matrix-based accuracy of 98.05\% and a Levenshtein distance-based accuracy of 87.20\%. Comparing our results with Tesseract on Bangla OCR primarily focused on handwritten data, our model demonstrates competitive performance across all document types.

Table ~\ref{table2} shows a comparison of our model's accuracy based on Levenshtein distance with tesseract in Bangla OCR. We were searching for available Bangla OCR to compare the performance. Unfortunately, there is no accessible, open-source Bangla OCR presented yet. Some of the commercial Bangla OCRs are available online. These results highlight the effectiveness of our model in accurately recognizing and extracting text from various types of Bengali documents.

\begin{table}[htbp]
\caption{Comparison of our model and tesseract ocr for Bangla documents}
\begin{center}

\begin{tabular}{|p{3cm}|p{1.5cm}|p{1.5cm}|}
\hline
\textbf{Document Type} & \textbf{Proposed Model} & \textbf{Tesseract}\\
\hline
Computer compose & 90.06\% & 73.49\%\\
\hline
Letterpress& 88.53\% &84.16\%\\
\hline
Typewriter & 83.38\% & 36.78\%\\
\hline
Handwritten & 86.84\% & 35.80\%\\
\hline
Average& 87.20\% & 57.56\%\\
\hline
\end{tabular}
\label{table2}
\end{center}
\end{table}

Specifically, the presented character recognition model \cite{charactermodel} gives an average recognition accuracy of 91.88\%, which is trained on 1,069,132 images having 207 root characters, ten modifiers, and six consonant diacritic classes. The model shows an individual recognition accuracy of 86.09\% for root characters, 95.56\% for modifiers, and 93.99\% for consonant diacritic. Table \ref{tablecomp} compared our word model accuracy with other Bangla OCRs.

\begin{table}[htbp]
\caption{Comparison of our word model vs other Bangla OCR}
\begin{center}
\begin{tabular}{|p{6cm}|p{1.5cm}|p{1.5cm}|}
\hline
\textbf{Model}  & \textbf{Accuracy}\\
\hline
Multilayer Perceptron \cite{isthiaq2020ocr} & 71\%\\
\hline
Back Propagation Neural Network \cite{ahmed2013enhancing} & 89\%\\
\hline
Deep Convolutional Neural Network \cite{purkaystha2017bengali} & 91\%\\
\hline
Multi-headed CNN \cite{charactermodel} & 91\%\\
\hline
Multiclass CNN \cite{lay1} & 93\%\\
\hline
Our word model & 95\%\\
\hline
\end{tabular}
\label{tablecomp}
\end{center}
\end{table}

Our model exhibits strong performance, achieving high accuracy in different document types and outperforming existing approaches in Bangla OCR. The combination of our attentional multi-headed convolutional neural network and the various processing steps in the inference pipeline contributes to the superior performance of our OCR system.

\section{Conclusion}
In conclusion, our research paper presents a comprehensive and innovative Bengali OCR system with unique layout reconstruction and capabilities for the Bangla language. The system excels in reconstructing document layouts and preserving structures and alignment, including paragraphs, tables, images(graphics element, logo, signature), and numbered lists. It goes beyond restoring embedded images, ensuring fidelity to the original content. Advanced image and signature detection further enhance its accuracy and versatility. Specialized models for word segmentation cater to various document types, and our OCR system can handle static and dynamic handwritten inputs. It successfully recognizes compound characters in Bengali, ensuring comprehensive character recognition. The system's technical components optimize character and word recognition, leveraging techniques such as automatic skew and perspective correction, noise removal, and self-attentional neural networks. Additionally, the system integrates queuing mechanisms for efficient and scalable processing, facilitating handling large files. These contributions make our Bengali OCR system an effective solution for efficient and accurate text extraction and analysis in Bengali documents.

\section*{Acknowledgment}
The authors want to acknowledge the support and funding provided by the Enhancement of Bangla Language in ICT through Research \& Development (EBLICT) project under the Ministry of ICT, Government of the People's Republic of Bangladesh. This work was conducted at the Apurba-DIU Research Lab (ADRL), and the authors thank them for their support and resources in carrying out this research.

{\small
\bibliographystyle{unsrt}
\bibliography{egbib}

\begin{thebibliography}{10}

\bibitem{charactermodel}
AKM Shahariar~Azad Rabby, Md.~Majedul Islam, Nazmul Hasan, Jebun Nahar, and Fuad Rahman.
\newblock Borno: Bangla handwritten character recognition using a multiclass convolutional neural network.
\newblock In Kohei Arai, Supriya Kapoor, and Rahul Bhatia, editors, {\em Proceedings of the Future Technologies Conference (FTC) 2020, Volume 1}, pages 457--472, Cham, 2021. Springer International Publishing.

\bibitem{islam2021text}
Md~Majedul Islam, AKM Shahariar~Azad Rabby, Nazmul Hasan, Jebun Nahar, and Fuad Rahman.
\newblock Text-type extraction using a deep learning solution at the character level.
\newblock In {\em Soft Computing and Signal Processing: Proceedings of 3rd ICSCSP 2020, Volume 1}, pages 253--262. Springer Singapore, 2021.

\bibitem{haque2018onkogan}
Sadeka Haque, Shammi~Akter Shahinoor, AKM Shahariar~Azad Rabby, Sheikh Abujar, and Syed~Akhter Hossain.
\newblock Onkogan: Bangla handwritten digit generation with deep convolutional generative adversarial networks.
\newblock In {\em International Conference on Recent Trends in Image Processing and Pattern Recognition}, pages 108--117. Springer, Singapore, 2018.

\bibitem{omee2012complete}
Farjana~Yeasmin Omee, Shiam~Shabbir Himel, and Md~Abu~Naser Bikas.
\newblock A complete workflow for development of bangla ocr.
\newblock {\em arXiv preprint arXiv:1204.1198}, 2012.

\bibitem{pramanik2018shape}
Rahul Pramanik and Soumen Bag.
\newblock Shape decomposition-based handwritten compound character recognition for bangla ocr.
\newblock {\em Journal of Visual Communication and Image Representation}, 50:123--134, 2018.

\bibitem{rabby2018universal}
AKM Shahariar~Azad Rabby, Sadeka Haque, Shammi~Akther Shahinoor, Sheikh Abujar, and Syed~Akhter Hossain.
\newblock A universal way to collect and process handwritten data for any language.
\newblock {\em Procedia computer science}, 143:502--509, 2018.

\bibitem{TessOverview}
Ray Smith.
\newblock An overview of the tesseract ocr engine.
\newblock In {\em ICDAR '07: Proceedings of the Ninth International Conference on Document Analysis and Recognition}, pages 629--633, Washington, DC, USA, 2007. IEEE Computer Society.

\bibitem{Breuel2008TheOO}
Thomas~M. Breuel.
\newblock The ocropus open source ocr system.
\newblock In {\em Electronic imaging}, 2008.

\bibitem{china}
Qiang Huo, Yong Ge, and Zhi-Dan Feng.
\newblock High performance chinese ocr based on gabor features, discriminative feature extraction and model training.
\newblock In {\em 2001 IEEE International Conference on Acoustics, Speech, and Signal Processing. Proceedings (Cat. No.01CH37221)}, volume~3, pages 1517--1520 vol.3, 2001.

\bibitem{arab}
A.~Cheung, M.~Bennamoun, and N.W. Bergmann.
\newblock An arabic optical character recognition system using recognition-based segmentation.
\newblock {\em Pattern Recognition}, 34(2):215--233, 2001.

\bibitem{rabby2018ekush}
AKM Shahariar~Azad Rabby, Sadeka Haque, Md~Sanzidul Islam, Sheikh Abujar, and Syed~Akhter Hossain.
\newblock Ekush: A multipurpose and multitype comprehensive database for online off-line bangla handwritten characters.
\newblock In {\em International Conference on Recent Trends in Image Processing and Pattern Recognition}, pages 149--158. Springer, Singapore, 2018.

\bibitem{alam2010complete}
Md~Mahbub Alam and M~Abul Kashem.
\newblock A complete bangla ocr system for printed characters.
\newblock {\em JCIT}, 1(01):30--35, 2010.

\bibitem{bag2013survey}
Soumen Bag and Gaurav Harit.
\newblock A survey on optical character recognition for bangla and devanagari scripts.
\newblock {\em Sadhana}, 38:133--168, 2013.

\bibitem{silberpfennig2015improving}
Adi Silberpfennig, Lior Wolf, Nachum Dershowitz, Seraogi Bhagesh, and Bidyut~B Chaudhuri.
\newblock Improving ocr for an under-resourced script using unsupervised word-spotting.
\newblock In {\em 2015 13th International Conference on Document Analysis and Recognition (ICDAR)}, pages 706--710. IEEE, 2015.

\bibitem{jannatul2021matrivasha}
Jannatul Ferdous, Suvrajit Karmaker, AKM Shahariar~Azad Rabby, and Syed~Akhter Hossain.
\newblock Matrivasha: A multipurpose comprehensive database for bangla handwritten compound characters.
\newblock In Jo{\~a}o Manuel R.~S. Tavares, Satyajit Chakrabarti, Abhishek Bhattacharya, and Sujata Ghatak, editors, {\em Emerging Technologies in Data Mining and Information Security}, pages 813--821, Singapore, 2021. Springer Singapore.

\bibitem{ban1}
Tasnim Ahmed, Md.~Nishat Raihan, Rafsanjany Kushol, and Md~Sirajus Salekin.
\newblock A complete bangla optical character recognition system: An effective approach.
\newblock In {\em 2019 22nd International Conference on Computer and Information Technology (ICCIT)}, pages 1--7, 2019.

\bibitem{ban2}
Md~Rakibul Hasan, Anamika~Basak Pew, Sanzida Alam, Nafisa~Tasnim Rifha, Shamin~Yeaser Shams, Farhan Shahriar, and Rashedur~M. Rahman.
\newblock Smart ocr for recognizing bangla characters with craft and deep learning models.
\newblock In {\em 2022 IEEE 13th Annual Ubiquitous Computing, Electronics \& Mobile Communication Conference (UEMCON)}, pages 0573--0577, 2022.

\bibitem{lay1}
Md.~Majedul Islam, Avishek Das, Ibna Kowsar, AKM Shahariar~Azad Rabby, Nazmul Hasan, and Fuad Rahman.
\newblock Towards building a bangla text recognition solution with a multi-headed cnn architecture.
\newblock In {\em 2021 IEEE International Conference on Big Data (Big Data)}, pages 1061--1067, 2021.

\bibitem{lay2}
Liang Qiao, Hui Jiang, Ying Chen, Can Li, Pengfei Li, Zaisheng Li, Baorui Zou, Dashan Guo, Yingda Xu, Yunlu Xu, Zhanzhan Cheng, and Yi~Niu.
\newblock Davarocr: A toolbox for ocr and multi-modal document understanding.
\newblock In {\em Proceedings of the 30th ACM International Conference on Multimedia}, MM '22, page 7355–7358, New York, NY, USA, 2022. Association for Computing Machinery.

\bibitem{doctr}
Mindee.
\newblock doctr: Document text recognition.
\newblock \url{https://github.com/mindee/doctr}, 2021.

\bibitem{docbed}
Wenzhen Zhu, Negin Sokhandan, Guang Yang, Sujitha Martin, and Suchitra Sathyanarayana.
\newblock Docbed: A multi-stage ocr solution for documents with complex layouts.
\newblock {\em Proceedings of the AAAI Conference on Artificial Intelligence}, 36(11):12643--12649, Jun. 2022.

\bibitem{docperse}
Johannes Rausch, Octavio Martinez, Fabian Bissig, Ce~Zhang, and Stefan Feuerriegel.
\newblock Docparser: Hierarchical document structure parsing from renderings.
\newblock {\em Proceedings of the AAAI Conference on Artificial Intelligence}, 35(5):4328--4338, May 2021.

\bibitem{table1}
Zewen Chi, Heyan Huang, Heng{-}Da Xu, Houjin Yu, Wanxuan Yin, and Xianling Mao.
\newblock Complicated table structure recognition.
\newblock {\em CoRR}, abs/1908.04729, 2019.

\bibitem{kafka}
Matthias~J. Sax.
\newblock {\em Apache Kafka}, pages 1--8.
\newblock Springer International Publishing, Cham, 2018.

\bibitem{zookeeperoverview}
{Apache Software Foundation}.
\newblock Zookeeper.
\newblock \url{http://zookeeper.apache.org}, 2011.
\newblock (Accessed on 10/25/2023).

\bibitem{onnx}
Onnx: Open neural network exchange.
\newblock \url{https://onnx.ai}.
\newblock (Accessed on 10/25/2023).

\bibitem{triton}
{NVIDIA Corporation}.
\newblock Triton inference server | nvidia developer.
\newblock \url{https://developer.nvidia.com/triton-inference-server}.
\newblock (Accessed on 10/25/2023).

\bibitem{laven}
Li~Yujian and Liu Bo.
\newblock A normalized levenshtein distance metric.
\newblock {\em IEEE Transactions on Pattern Analysis and Machine Intelligence}, 29(6):1091--1095, 2007.

\bibitem{matrix}
Martin Reynaert.
\newblock Character confusion versus focus word-based correction of spelling and ocr variants in corpora.
\newblock {\em IJDAR}, 14:173--187, 06 2011.

\bibitem{Alberti}
Michele Alberti, Manuel Bouillon, Rolf Ingold, and Marcus Liwicki.
\newblock Open evaluation tool for layout analysis of document images.
\newblock In {\em 2017 14th IAPR International Conference on Document Analysis and Recognition (ICDAR)}, volume~04, pages 43--47, 2017.

\bibitem{isthiaq2020ocr}
Asif Isthiaq and Najoa~Asreen Saif.
\newblock Ocr for printed bangla characters using neural network.
\newblock {\em International Journal of Modern Education and Computer Science}, 12(2):19, 2020.

\bibitem{ahmed2013enhancing}
Shamim Ahmed and Mohammod~Abul Kashem.
\newblock Enhancing the character segmentation accuracy of bangla ocr using bpnn.
\newblock {\em International Journal of Science and Research (IJSR) ISSN (Online)}, pages 2319--7064, 2013.

\bibitem{purkaystha2017bengali}
Bishwajit Purkaystha, Tapos Datta, and Md~Saiful Islam.
\newblock Bengali handwritten character recognition using deep convolutional neural network.
\newblock In {\em 2017 20th International conference of computer and information technology (ICCIT)}, pages 1--5. IEEE, 2017.

\end{thebibliography}
}

\end{document}